\documentclass[a4paper]{article}

\usepackage{INTERSPEECH2021}

\def\onlyINTERSPEECH#1{#1}

\def\ifinterspeech#1{#1}
\def\ifiwslt#1{}

\def\citet#1{\cite{#1}}

\title{%Speech Translation from Source or Interpreter?
Lost in Interpreting: Speech Translation from Source or Interpreter?
}
\onlyINTERSPEECH{\name{Dominik Mach{\' a}{\v c}ek, Mat\'{u}\v{s} \v{Z}ilinec, Ond{\v r}ej Bojar%
\thanks{The research was partially supported by the grants 19-26934X  (NEUREM3)  of  the  Czech  Science Foundation, H2020-ICT-2018-2-825460 (ELITR) of the EU, and 398120 of the Grant Agency of Charles University.
}}
\address{
  Charles University, 
  Faculty of Mathematics and Physics \\%
  Institute of Formal and Applied Linguistics
 }
\email{\{surname\}@ufal.mff.cuni.cz}
}

\usepackage{graphicx}
\usepackage{xcolor, colortbl}
\usepackage{cleveref}
\usepackage{multirow}
\usepackage{rotating}
\usepackage{url}

\usepackage[normalem]{ulem} % for sout

\def\XXX#1{{\textcolor{red}{XXX #1}}}
\def\hideXXX#1{}
\def\XXXifaccepted#1{}
\def\XXXifaccepted#1{\XXX{#1}}

\def\furl#1{\footnote{\url{#1}}}
\def\captiongap{\vspace{-3mm}}

\begin{document}

\maketitle
\begin{abstract}
Interpreters facilitate multi-lingual meetings but the affordable set of languages is often smaller than what is needed.
Automatic simultaneous speech translation can extend the set of provided languages. We investigate if such an automatic system should rather follow the original speaker,
or an interpreter to achieve better translation quality at the cost of increased
delay.

To answer the question, we release Europarl Simultaneous Interpreting Corpus (ESIC), 10 hours of
recordings and transcripts of European Parliament speeches in English, with
simultaneous interpreting into Czech and German. We evaluate quality and latency of
speaker-based and interpreter-based spoken translation systems from English to
Czech.
We study the differences in implicit
simplification and summarization of the human interpreter compared to a machine
translation system trained to shorten the output to some extent. Finally, we perform human evaluation to measure information
loss of each of these approaches.
%\XXX{old:
%Simultaneous speech translation can be used to increase the language coverage of
%live events. This can be performed in two ways: either by directly using the
%original speech as the source, or indirectly, translating the speech of a human
%interpreter from an intermediate language. In this work, we investigate the
%features of these two options.
%in order to answer the question when it is more suitable to use one of these over the other. 
%}
\end{abstract}

\ifinterspeech{
\noindent\textbf{Index Terms}: speech translation, machine translation, %speech translation corpus, 
simultaneous interpreting corpus, interpreting
}

\section{Introduction}

%XXX{Před odesláním na IWSTL:
%1) jako hlavní soubor v menu overleafu nastavit main-iwslt.tex, to použije tu %správnou šablonu
%\sout{2) opravit závorkování citací (v interspeechi to nebylo)}
%\sout{3) dát popisky tabulek pod ně}
%4) zkontrolovat délku 4-8 stránek + reference
%}

Multilingual events with participants without a common language are often
simultaneously interpreted by humans.
Automatic simultaneous speech translation can increase the language coverage
%of
%live events. It is especially useful
%for languages in which
where human %simultaneous
interpreting is not available, e.g. because of capacity reasons.
Assuming the presence of a human interpreter, speech translation can
%be performed
%in two ways:
%either by directly
%by using 
rely on
the original speech as the source, or by
translating the speech of the interpreter. In this work, we compare the features of these two options. 

The direct source-to-target translation is supposed to be fast (no
latency introduced by the interpreter), and more literal,
%presumably word-for-word,
and therefore very detailed. However, the
verbosity might be uncomfortable for final users to follow,
if the speech is too fast or disfluent. The indirect interpreter-to-target
translation might benefit from the fact that interpreters tend to compress
and simplify \cite{he-etal-2016-interpretese,eptic}, on the other hand, it could
decrease adequacy. 

In this work, we examine two possible sources and one target language. However, we put aside the effects of varying quality of speech recognition and machine translation. They
can favor any option, depending on the specific version of the tools and other conditions. We focus on the evaluation of latency, shortening and simplification, and
human assessment of information loss. We prepare a new evaluation corpus ESIC (Europarl Simultaneous Interpreting Corpus v.1.0) %from European Parliament, 
with 10 hours of
English speeches with transcripts, translations and transcripts of simultaneous interpreting into Czech and German. 

%XXX{the findings, very briefly. Pokud se vejdou a stihnou.}

%On one hand, translating the interpreter adds latency to the system. On the other hand, the simultaneous interpreter implicitly summarizes and simplifies the speech \cite{he-etal-2016-interpretese}. Hence, translation of the interpreted speech might be more comfortable for reading. Direct automatic translation might then more accurately preserve information from the original speaker while still being easy to receive, especially if the model is additionally biased to shorten the source text. Additionally, the interpreter provides inter-cultural transfer and organizational comments, which might not present in the original speech. In this work, we aim to investigate the advantages of these two options.

%We put aside the effect of language similarity and complexity of translation between distinct language pairs. 

\section{Related Work}

The plenary sessions of European Parliament (EP) are a useful source of parallel
data, %. They are already processed in 
known well from
the  multi-parallel text-to-text corpus
Europarl \cite{koehn2005epc}. The recent speech-to-text corpus Europarl-ST
\cite{jairsan2020a} is a collection of short audio-translation segments for
bilingual or multi-target speech-to-text translation. It contains only the
audio of original speakers, not the interpreters.

The corpora EPTIC \cite{eptic}, EPIC \cite{sandrelli-bendazzoli-2006-tagging}
and EPIC-Ghent \cite{epic-ghent} are small collections of transcribed
interpretings from European Parliament %. They were
created for analyses of
interpreting. They contain only selected languages, not including English,
German and Czech. They do not contain timestamps and audios of interpreting, and
their accessibility is restricted. The other corpora of simultaneous
interpreting \cite{temnikova-etal-2017-interpreting,pan-2019-chinese} focus on other languages.

Additionally, text simplification in the context of machine translation remains an open problem. The existing methods focus on augmenting the translation model with length tokens or positional encoding to control the length of the output text \cite{kikuchi-etal-2016-controlling,takase-okazaki-2019-positional}. For an overview, we refer the reader to \ifinterspeech{Lakew }\citet{lakew}.
% Another approach is by pipelining MT + summarization model. [but] Abstractive summarization has been studied mostly in the news context - generating headlines from a sentence (\XXX{tato citacia je trochu random!}) \cite{chopra-etal-2016-abstractive}. 
% Neviem, ci by fungovalo pouzit abstractive summarization a ci to zmienovat, pretoze to produkuje skor titulky pre noviny ako suvisly text ...

\section{ESIC: Corpus Composition}

Since 2008, the EP is publishing the audios of simultaneous interpreting into
all 22 EU official languages in that time. Until 2011, it was publishing the revised
transcripts and translations into all EU languages. The period of 2008 to 2011
is a valuable resource containing parallel revised translations and simultaneous
interpreting, which we decided to study.
%The other periods are left for future work.

We focus on English%
%as a source language,
%the most common European lingua
%franca. We downloaded and processed
, the most common European lingua franca, as the source, and on
simultaneous interpreting into German and
Czech. German is a language with second most speakers in EU, and it often serves as interpreting target at
%might be interpreted into it on 
many international events. Czech is an example
target language into which it might be translated automatically.

We downloaded the data and aligned the revised transcripts and audio by
metadata. We processed the speeches with automatic diarization
\cite{Rouvier13anopen-source} to roughly annotate their beginning and end
timestamps in long recordings of the whole sessions.
For simplicity, we decided to exclude the president because his or her
utterances while chairing the sessions were often not transcribed, or not
word-for-word. We also excluded speeches which we could not align due to error
in metadata or in automatic processing, which were shorter than 30 seconds, or whose Czech translation or interpreting was missing. 

Next, we selected 10 hours of speeches into validation and evaluation set. We
decided to eliminate the potentially malicious overlap of ESIC dev-test with
Europarl-ST train set. 
We identified the speakers of Europarl-ST English-German dev-test, found all their speeches in our data, and included them into ESIC dev-test. 
%\XXX{Nesrozumitelne. Doporucuju rikat osoby: "if..., we exclude...": Taky mam pocit, ze je tu zameneny nekde Europarl-ST a ESIC:}
%If a speech is in Europarl-ST English-German dev-test, or
%the given speaker is included in Europarl-ST English-German dev-test, then we do not include it in ESIC.
% mozna to Tvoje je spravne formulovana implikace, ale ctenar nedokaze delat mnozinove vypocty, kdyz ty mnoziny nema ukotvene v pameti.
To cover full 10 hours, we added additional 28 randomly selected speeches,
regardless the speakers in Europarl-ST. We marked them so that the users can be aware.
%We mark this set as dev2. \XXX{pozor, dev2 neni v tabulce!}

%\vspace{-3mm}%
\subsection{Manual Revisions}

\def\rev{Revised}
\def\orto{Ortho}
\def\verbatim{Verbatim}
\def\duration{Duration}

We manually revised the segmentation into individual speeches in all three
tracks (English source, Czech and German interpreting) because the automatic
diarization was inaccurate at beginnings and ends. 
In the next steps, we manually transcribed the interpreters following fixed annotation
guidelines. Our annotators marked false starts, unintelligible words, short
insertions in different languages and swapping voices, so that ESIC users can
decide to handle them in a particular way. They transcribed and marked the segments which
could not be easily transferred from orthography to verbatim, e.g. the
non-canonical forms of numerals, dates, loaned named entities and acronyms. They
inserted orthographic punctuation and spelling, but did not do any changes in
syntax, even when the interpreter's syntax could be considered as ungrammatical.
Hesitations were not marked.
In sum, we ended up with three versions: \rev{} as downloaded from the web, \verbatim{} which does not include any punctuation, but does include false starts, and \orto{} with punctuation and without false starts.
%\XXX{confusing wording: Ortho with correct Verbatim}correct \verbatim{} in the orthography but not syntax.

The transcripts of English sources were revised in the same way as those of interpreters',
but the annotator re-used the transcripts from the web, which were manually
revised and normalized by EP staff for comfortable reading. They often differed
from the verbose ones in the way of addressing the president and Parliament at the
introduction, in the correction of disfluencies and grammar, use of more formal named
entities or decompressed acronyms, and removal of side and organizational
comments. Also, the concluding ``thank you'' to the president was added by our
revision.

Furthermore, our annotator marked, with the use of the video-recording, whether the speech was
spontaneous, or read, because we believe it has a big impact on the grammar,
style and complexity of translation.
%, and this information should be considered in analyses. % OB: sami to nedelame, tak nepritahovat pozornost
%
In rare cases, we excluded speeches given in another language
than in English, but short code switching, e.g. the
salutation of the president in his or her native language, were kept for
authenticity.

%\subsection{Automatic Postprocessing}

Finally, we used MAUS forced aligner \cite{kisler2017multilingual} for English,
German and Czech to obtain the word-based timestamps. 
The corpus statistics are in \Cref{tab:esic-stats}.

\def\tabstatcaption{
        \caption{Size statistics of ESIC corpus. The two numbers in each
		cell are the number of sentences (or documents, in the row of \verbatim{} transcription), and number of words.}
        \label{tab:esic-stats}
}
\def\twocol#1{\multicolumn{2}{r|}{#1}}
\def\lasttwocol#1{\multicolumn{2}{r}{#1}}
\begin{table}[]
    \centering
    \ifinterspeech{\tabstatcaption{}\captiongap{}}
    \footnotesize{}
    \begin{tabular}{@{}l@{~}l|r@{~~}r|r@{~~}r|r@{~~}r}
	% Note: Following "dev" is dev+dev2
~                       &          ~  &           \multicolumn{2}{c|}{\bf Source}  &  \multicolumn{4}{c}{\bf  Interpreting into}\\
~                       &          ~  &           \multicolumn{2}{c|}{\bf English}  &  \multicolumn{2}{c|}{\bf  German} &      \multicolumn{2}{c}{\bf  Czech}\\
\hline
\multirow{4}{*}{Dev}&   \rev       &  2019              &  44986             &  2015                   &   42969  &  2019  &  37017  \\
&                       \verbatim  &  179               &  47478             &  179                    &   38956  &  179   &  33863  \\
&                       \orto      &  2772              &  45862             &  2818                   &   38482  &  2736  &  33163  \\
&                       \duration  &  \twocol{5h8m38s}  &  \twocol{5h9m17s}  &  \lasttwocol{5h10m30s}  \\
\hline
\multirow{4}{*}{Test}&  \rev       &  1997              &  45068             &  1991                   &   42347  &  1997  &  36600  \\
&                       \verbatim  &  191               &  47331             &  191                    &   39115  &  191   &  34464  \\
&                       \orto      &  2693              &  45640             &  2900                   &   38738  &  2720  &  33747  \\
&                       \duration  &  \twocol{5h3m54s}  &  \twocol{5h2m23s}  &  \lasttwocol{5h6m16s}   \\
    \end{tabular}
    \ifiwslt{\tabstatcaption{}}
    \ifinterspeech{\vspace{-2mm}}
\end{table}

\subsection{Ethics}

We received the authorisation to repackage and publish the texts and audios of the
speakers on the EP plenary sessions, and the transcripts of interpreters\footnote{Available at \url{http://hdl.handle.net/11234/1-3719}.}. Since
the interpreters' voices are considered as personal data, we do not publish them
together with the corpus. However, they are publicly available on the web of EP,
and we can publish the links and instructions that every user of our corpus can
follow to obtain them.

%The corpus can be accessed by contacting the authors. If accepted, the link will
%be included.

\section{Translation Systems}

% makra pro jazykove pary (a MT systemy)
\def\encs{\textsc{en-cs}}
\def\decs{\textsc{de-cs}}
\def\csint{\textsc{cs-int}}
\def\csref{\textsc{cs-ref}}
\def\deint{\textsc{de-int}}
\def\enasr{\textsc{en~ASR}}
\def\deasr{\textsc{de~ASR}}

In the next sections, we compare three options for translation of English speech into Czech: human
interpreting into Czech (\csint{}), human interpreting into German (\deint{}) followed by a machine
translation system into Czech (\decs{}), and a machine translation model directly into Czech, 
which was additionally trained to shorten source text (\encs{}). 

\vspace{-2mm}%
\subsection{Machine Translation}

\encs{} is a Transformer-Base \cite{attention-is-all-you-need:2017} machine translation model trained using
Marian \cite{mariannmt} on CzEng 1.7 %\XXX{existuje 1.7 clanok? nenasiel som, citujem 1.6} % OB: spravne, 1.7 je nejlepe popsana v 1.6
\cite{czeng16:2016} using the default hyperparameters. It
was  biased during training by providing training examples
illustrating shortening. %where the target side is shorter than the source side.
Specifically, sentence pairs from the parallel corpus were selected only if the Czech sentence had not more than 86\% of the number of subword units compared to the English counterpart. Given that in the CzEng corpus, Czech sentences are on average 10\% longer than their English translations in terms of subword units, our requirement corresponds to \textsc{en}:\textsc{cs} compression factor of 1:0.78. %(English:Czech).

In comparison to an identical model trained on the full corpus, we observed a decrease in both mean length of the translation and BLEU score with the shortening model.

Furthermore, we observed that the model often performs shortening by replacing words and phrases with their
synonyms  with fewer subword units, but preserves the syntax, which does not significantly differ from the baseline non-shortening model's translation. 
This is in contrast to human interpreting strategy \cite{he-etal-2016-interpretese}. Human interpreters tend to segment the source sentence into small units and translate them as individual sentences. Furthermore, they use generalization and summarization of the whole clauses, and other techniques such as passivization to consolidate the word order between source and target.

% {copy-pastuju ze Subtitler User Study:}
\decs{} is trained on 8M sentence pairs from Europarl and Open
Subtitles \cite{koehn2005epc,lison-tiedemann-2016-opensubtitles2016}, the
only public parallel corpora of German and Czech, and validated on newstest.
The Transformer-based system runs in Marian \cite{mariannmt} and reaches 18.8 cased BLEU on WMT newstest-2019.
%\XXXifaccepted{
%It was described in \cite{kvapilikova-etal-2019-cuni} as the supervised
%benchmark.
It is not adapted for simultaneous translation which would need translation stability and partial translation for partial sentences \cite{Niehues2018,Arivazhagan2020ReTranslationSF}.
%}

\vspace{-2mm}
\subsection{Low-Latency ASR}
% Ahoj. Prosim koukni do ACK, potreboval bych vymaznout SVV, muzeme?
% a take prosim neblokovat globalne XXX, naopak vyresit/zrusit vsechny jeho vyskyty

We use online German and English ASR systems originally prepared for lectures
\cite{cho-real-world}. They emit partial hypotheses
in real time, and correct them as more context is available. German is a hybrid HMM-DNN model (\deasr{}). The same system
was used also by KIT Lecture Translator \cite{muller-etal-2016-lecture}. English is neural sequence-to-sequence ASR \cite{nguyen2021superhuman}. 
They are connected in a cascade with a tool for removing disfluencies and
inserting punctuation \cite{Cho2012SegmentationAP} and with the MT systems.
The cascade is the same as the one of the ELITR project at IWSLT 2020
\cite{machacek-etal-2020-elitr}.

%\subsection{Sources}

%\XXX{DM: tahle podsekce je divná... To sem nepatří, ne?}

%As the source speech, we use the original English audio recordings of EP speakers from the ESIC test set. 
%Additionally, in evaluation we use the exact reference translation into Czech (\csref{}).

\def\rot#1{
\begin{sideways}
#1
\end{sideways}
}
\def\p{$\pm$}

\section{Latency}

% Co tam potřebuju napsat:
% - jednu souhrnou závěrečnou tabulku
% - vysvětlit finalizaci MT -- že je to spíš horní odhad. Docela přísná finalizace.
%    - vysvětlit, že SOTA by to uměl líp. Něco, co 
% - může se tam dát graf zpoždění u jedné promluvy -- z kterýho jsou vidět TU?
%      - dát do souvislosti s hranicema vět?
%      - nápad: najdeme TU zdroje podle tlumočníků, pošleme je do MT -> co to udělá s latencí a kvalitou? 
%         - nešlo by použít online tlumočníka na naučení segmentace zdroje?
% - vysvětlit alignment src-tgt
% - Conclusion -- 

%%%%%%%%%%%%%%%%%%%%%%%%%%%%%%%%%%%%%%%%%%%%%%%%%%%%%%%%%%%%%%%%%%%%%%%%%%%%%%%%%%%%%%%%%%%%%%%%%%%%%%%%%%%%%%%%%%%%%%%%%%%%%%%%%%%%%%%%%%%%%%%%%%%%%%%%%%%%
% Latency -- přepsáno odznova

We aim to compare the latency of interpreting and machine translation. Note that the comparison is inevitably limited by different output modalities. The interpreters produce speech, and the machine translation text. We disregard the perception effects of hearing versus reading.

%We define the ``latency'' as the difference of times of the source word and its corresponding word in the target (interpreting, or machine translation).  
We need to assess the time when each word in source, interpreting and machine translation was produced.
For the source and interpreting, we have word-based timestamps from forced alignment tool.
For the re-translating machine translation, we use the finalization time of a target word as in \citet{Arivazhagan2020ReTranslationSF}. It is the first time when the system produces the word, and the word and all its preceding words remain unchanged until the end of the session.
This definition is rather harsh because it penalizes subtle, cosmetic changes in translation output the same way as meaning-altering re-translations. It is possible that a real user reads the translation earlier than at finalization time, and does not notice short flicker in previous words. However, the finalization time is an upper bound for the word production time. 

The ``latency'' is the difference of times of the source word and its ``corresponding'' word in the target. We assess the correspondence with automatic word alignment.

%\vspace{-2mm}%
\subsection{Word Alignment}

%We use the automatic word alignment tool fast\_align \cite{dyer-etal-2013-simple} to align English source transcripts and target interpreting or machine translation. 
We aligned English source transcripts and target interpreting or machine translation at the word level with fast\_align \cite{dyer-etal-2013-simple} after tokenizing
%We tokenized the texts
\cite{koehn-etal-2007-moses} and trimming them to 5 characters as a trivial form of lemmatization. We processed all 370 ESIC documents, treating each as a single sentence. We added relevant sentence-aligned texts to fast\_align training data, to expand the vocabulary: revised translations of Europarl 
(around 4 thousands documents from the same period) for interpreting, and the source and target sentence prefixes for machine translation.
We obtained forward and backward alignments, and removed those going back in time, assuming that the interpreters do not risk predicting content. Finally, we intersected them. 
Based on a small manual check, the resulting word alignments were reasonably good, despite that fast\_align is designed for individual sentences and our documents were much longer. 
%Occasionally, there were mis-alignments of tokens appearing multiple times in the document. However, we assume that their frequency is the same across the translation candidates, and they will be lost in a

%\vspace{-2mm}%
\subsection{Latency Comparison}

The latency is summarized in \Cref{tab:latency}. Both \csint{} and \deint{} have average latency around 4 seconds. In 90\% of the source words that were aligned to any target word, the latency is below 7 seconds. In small number of cases, in around 1\%, the latency is larger than 23 seconds. It can be caused either by interpreters using so long translation unit, or a rare error in the automatic alignment. The methodology is the same for all options, therefore we assume that the error rate is homogeneous, although unknown, so the results are comparable.

The machine translation systems used in our work have larger latency than interpreters: \encs{} around 7 seconds, \decs{} around 5 seconds. There are two reasons why their latencies differ, and why they are so large. First, \encs{} uses end-to-end ASR, which is approximately 1 second slower than the hybrid ASR of \decs{}. Second, both systems are used for re-translating growing system prefixes, despite they were trained on full sentences. The first word in the sentence is often finalized after the whole sentence is completed by the speaker. The English source speakers tend to make long sentences, sometimes even 30 seconds, while the \deint{} makes shorter ones. 

The systems thus translate much longer units than interpreters, and therefore have larger latency. 
We hypothesize that more advanced translation system could have latency comparable to the interpreter. Assuming that the interpreters always wait optimally for meaningful translation units, their latency is an upper bound for the waiting. Machine processing (speech recognition and translation) can take up to 1 second. 
ESIC corpus can serve for tuning the parameter $k$ of wait-$k$ models \cite{ma-etal-2019-stacl} for simultaneous translation, so the resulting latency of wait-$k$ is the same as interpreters'.

%Machine processing takes around 1 second, so we assume that optimal simultaneous speech translation could reach 5 second latency not only in average, but also in at least 90\% of the source.

The indirect \deint{}+\decs{} option has latency around 10 seconds between English and Czech, i.e. roughly twice larger than a single interpreter. This is comparable to relay interpreting via one intermediate pivot language. Relay interpreting is used in real-life settings, so real users might be accustomed to latencies around 10 seconds. Therefore, we consider the indirect path of interpreter followed by machine translation as feasible from the latency point of view.

\def\tablatencycaption{
        \caption{Latency of interpreting and machine translation from English to Czech (white background), based on automatic word alignments, in seconds. Gray rows break down the two intermediate components of the indirect translation: English-to-German interpreter and German-to-Czech translation. 
        The percentile indicates that, e.g. 90\% of aligned words fit under 7 seconds. %The column ``short'' denotes the proportion of source and target tokens, ``al.'' denotes the percentage of aligned source tokens.
        %\XXX{Navrhoval bych prejit na vteriny a psat jen 2 des. mista: 4.15. Bylo by to o dost citelnejsi.
        %DM: OK. Jdu předělat teď. + navíc short a al se dá dát pryč
        %}
        }
    \label{tab:latency}
}

\definecolor{Gray}{gray}{0.9}
\begin{table}[]
    \centering
    \ifinterspeech{\tablatencycaption{}%
    \captiongap{}%
    \vspace{-1mm}
}
    \begin{tabular}%{ll|rrrrrrrr}
    {@{}l@{~~}l@{~~}|@{~~}r@{~~}|@{~~}r@{~~~}r@{~~~}r@{~~}}
    &   &                & \multicolumn{3}{c}{Percentile $\le$}  \\
 &  &     avg\p{}std & $50\%$ & $90\%$ & $99\%$ \\
   \hline
    
    \multirow{5}{*}{\rot{dev}} &
% DEV

\csint{} & % interpreter only csen dev? 
4.17 \p~4.32 & 3.21 & 7.06 & 22.14  \\ 

& \encs{} & % mt-encs dev? &
7.56 \p~5.65 & 5.97 & 15.26 & 27.00  \\ 

& \deint{}+\decs{} & % int+mt-encs-de dev? &
9.90 \p~6.75 & 8.57 & 17.00 & 34.78  \\ 

\rowcolor{Gray}
& ~~~(\deint{}) %interpreter only deen dev? &
&
4.26 \p~5.00 & 3.08 & 7.34 & 24.88  \\ 

\rowcolor{Gray}
& ~~~(\decs{}) & % mt-decs dev? &
4.92 \p~4.78 & 3.75 & 10.17 & 21.38 \\ 

\hline
%\hline

% TEST
\multirow{5}{*}{\rot{test}} &
\csint{} & % interpreter only csen test &
3.99 \p~4.38 & 3.00 & 6.77 & 22.23 \\ 

& \encs{} & % mt-encs test &
7.68 \p~6.28 & 5.98 & 15.17 & 30.38 \\ 

& \deint{}+\decs{} & % int+mt-encs-de test &
9.84 \p~7.16 & 8.43 & 17.08 & 36.70 \\ 

\rowcolor{Gray}
& ~~~(\deint{}) & % interpreter only deen test &
4.03 \p~4.70 & 3.02 & 6.64 & 23.27 \\ 

\rowcolor{Gray}
& ~~~(\decs{}) & % mt-decs test &
5.07 \p~4.89 & 3.90 & 10.56 & 20.95 \\ 
    \end{tabular}
    \ifiwslt{\tablatencycaption{}}
    \ifinterspeech{\vspace{-2.5mm}}
\end{table}

%%%%% konec Latency
%%%%%%%%%%%%%%%%%%%%%%%%%%%%%%%%%%%%%%%%%%%%%%%%%%%%%%%%%%%%%%%%%%%%%%%%%%%%%%%%%%%%%%%%%%%%%%%%%%%%%%%%%%%%%%%%%%%%%%%%%%%%%%%%%%%%%%%%%%%%%%%%%%%

\section{Shortening and Complexity}

%%%%%%%%%%%% Schováno Dominikem. Počet tokenů nic neříká.
%In \Cref{tab:shortratio}, we report length ratios between the translation and source text, in tokens, for our evaluated systems. Notably, all the systems achieve a shortening of the source text with a similar resulting length, but seem to do so using different methods. 
% \begin{table}[h!]
% \centering
% \caption{Length ratios of translation (second column) to reference (first column). }
% \label{tab:shortratio}
% \begin{tabular}{ll|r} \\ 
% Reference & System & Length ratio  \\ 
% \hline
% \csref{} & \csint{} & 90.04\ \% \\
% \csref{} & \decs{}  & 88.84\ \% \\
% \csref{} & \encs{}  & 87.98\ \% \\ 
% \hline
% \textsc{en-src} & \csint{} & 75.14\ \% \\
% \textsc{en-src} & \decs{}  & 74.11\ \% \\
% \textsc{en-src} & \encs{}  & 73.39\ \% \\
% \hline
% \textsc{en-src} & \csref{} & 84.17\ \% 
% \end{tabular}
% \end{table}

%  238754 cs.ref1
% 284869 cs.ref2
% 238844 decs.mt
% 270813 de.src
% 248845 encs.mt
% 269121 en.src

%Cca 90% pre vsetky 3.
%\XXX{TODO: Aky je nas zaver? Ze interpreter skracuje lepsie takze clovek moze rychlejsie citat, ale za cenu vyssej latencie?}

% přepsáno, do IWSLT:

We aim to compare the shortening and simplification capability of interpreting vs direct machine translation systems. 

First, the translation length. Syllables are units independent on the orthography and phonemic inventory of the languages, and they are capable to express shortening rate of translation into multiple languages.
Therefore, we used grapheme-to-phoneme and syllabification tool \cite{webmaus-g2p} for estimating the number of syllables in English, Czech and German source, interpreting and translation. 
%The tool internally uses tokenization, word normalization, grapheme-to-phoneme conversion and syllabification.  %% neni potřeba, zkracujeme
The results are in \Cref{tab:translen}. We also demonstrate that German uses more characters per syllable than Czech, due to smaller character inventory. This fact has to be considered especially in speech-to-text translation.
%While in Czech, there is almost 1:1 correspondence between phonemes and characters, in German and English
%The result

The results show that there is nearly no difference in translation length of interpreting, indirect \deint{}+\decs{}, and our shortening model for direct speech translation (\encs{}). On average, one English syllable is translated into one Czech syllable. The revised text translation \csref{} are longer than source, there is 1.19 syllable for 1 source syllable. The first reason might be that it is manually revised and adapted for reading. Shortening and simplification is not desirable in translation, while in interpreting it is necessary. The second possible reason is that interpreting might be unreliable. It may contain outages, and therefore be short.

%\subsection{Complexity of Vocabulary}

Next, we compare the vocabulary complexity. We rank Czech words from the CzEng corpus by frequencies, such that the most common word has rank 1, and the least common word has the rank of number of unique words. The ``comma'' and ``full stop'' characters were removed before the evaluation. \Cref{tab:rank} shows the mean and standard deviation of log ranks for each system across the documents in the test set.
We test whether the mean log rank of \encs{} is statistically equal to that of \decs{}. Using the two-sample Z-test, we reject this hypothesis with $p < 0.01$.
Thus, we conclude that the translations \encs{} (machine) and \csref{} (human), which do not contain any interpreter component, use a more complex vocabulary than both setups involving an interpreter, \csint{} and \decs{}. 

% \p
% \begin{table}[h!]
%   \centering
%         \caption{\XXX{}}
%         \label{tab:rank}
% \begin{tabular}{lrr}
% System  & Mean  & Stddev \\ \hline
% encs.mt & 9964  & 30277 \\
% decs.mt & 8578  & 27978 \\
% cs.ref1 & 8614  & 28227 \\
% cs.ref2 & 10112 & 31275 \\
% Mean word rank: 9317
% \end{tabular}
% \end{table}

%\subsection{Translation Length}

\def\tablatencycaption{\caption{Length rate of source to target of ESIC test set. For example, \csref{} has 1.19-times more syllables than English source. There is average and standard deviation on all test documents.}
\label{tab:translen}}
\begin{table}[]
    \centering
    \ifinterspeech{\tablatencycaption{}\captiongap{}}
    \begin{tabular}{l|r|rl}
System         &  Syllables & Characters \\
\hline
\csref{} & 1.19\ \p\ 0.12 & 0.93\ \p\ 0.09 \\
\csint{} & 1.03\ \p\ 0.17 & 0.80\ \p\ 0.13 \\
\encs{} & 1.03\ \p\ 0.10 & 0.82\ \p\ 0.04 \\
\deint{}+\decs{} & 1.01\ \p\ 0.16 & 0.79\ \p\ 0.12 \\
\hline
\deint{} & 1.01\ \p\ 0.15 & 0.99\ \p\ 0.14 \\
% syll de-all/de-gold-transcript 1.01 0.15
% chars de-all/de-gold-transcript 0.99 0.14
% words de-all/de-gold-transcript 0.86 0.13
    \end{tabular}
\ifiwslt{\tablatencycaption{}}
%\ifinterspeech{\vspace{-7mm}}
\end{table}

\def\tabrankcaption{
        \caption{Mean and standard deviation of log word frequency ranks calculated from translations of the test set. The column ``words" denotes the sample size (number of words in the translation). The proportion of out-of-vocabulary words is less than 0.5\ \% for each system.}
        \label{tab:rank}
}
\begin{table}[t]
  \centering
\ifinterspeech{\tabrankcaption{}\captiongap{}}
\begin{tabular}{l|r@{~}r@{~}ll}
System  & avg & \p & std & words \\ \hline %\hline
\encs{} & 6.42 & \p & 2.89 & 32\,488 \\
\decs{} & 6.16 & \p & 2.85 & 32\,703 \\
\csint{}  & 6.15 & \p & 2.83 & 32\,992 \\
\csref{}  & 6.32 & \p & 2.93 & 37\,182
\end{tabular}
\ifiwslt{\tabrankcaption{}}
%\ifinterspeech{\vspace{-2mm}}
\end{table}

\section{Quality}

We estimate the quality of machine translation with an automatic metric, and manually assess content preservation.
%Next, we quantify the difference in vocabulary complexity and content preservation of these two approaches to translation.

\vspace{-2mm}%
\subsection{BLEU against two References}

In \Cref{tab:bleu}, we provide the BLEU \cite{papineni-etal-2002-bleu} score of the indirect translation of German interpreting (\decs{}) and the direct \encs{} translation. We measure the score against two possible references: the revised text translation, and transcript of Czech interpreting. The sources are gold transcripts, not ASR, therefore it is an upper bound for translation quality in a real event.

We expected that \decs{} will be closer to \csint{} reference than \encs{}, but it is not. It might be caused by different interpreting strategies, and variability of translation, and too literal translation from German.  %We hypothesize that it could be caused by different interpreting strategies in both languages, 
%The BLEU score is lower for \decs{} in both cases. We assume it is because German 
%Our expectation was that the \encs{} system would perform better in terms of BLEU score against the reference, because it does not summarize, but lower score against interp
%
%The BLEU scores reflect this to some extent, where the \decs{} system receives a much lower score than \encs{} for the exact translation reference, but the difference is smaller for the Czech interpreter reference. Therefore, we conclude that machine translation (\encs{}) matches the source text more exactly than interpreter (\decs{}). 
We however refrain from the interpretation that \decs{} is of lower quality, since it has been previously shown that BLEU negatively correlates with simplicity \cite{sulem-etal-2018-bleu}.

\def\tabbleucaption{
        \caption{
        BLEU score between \encs{}, \decs{} and both Czech reference translations. 
        %The column ``BLEU agg'' denotes BLEU score calculated on the entire test set merged into one sequence. 
        %\hideXXX{better word for BLEU unit that's not line?}
        %The column ``BLEU one'' denotes BLEU score calculated on the test set by treating each individual speech recording as one \repl{line}{sequence}.
        BLEU requires a 1-1 correspondence between candidate and reference segments. We either treat the whole test set as one segment (``BLEU agg'') or each speech in the test set as one segment (``BLEU one'').
        }
        \label{tab:bleu}
}
\begin{table}[t]
  \centering
\ifinterspeech{\tabbleucaption\captiongap{}}
\begin{tabular}{l|l|cc}
Reference & System     & BLEU agg  & BLEU one  \\ \hline \hline
\csint{} & \encs{}  & 21.4 & 13.8  \\
\csint{} & \decs{}  & 19.9 & 10.4  \\ \hline
\csref{} & \encs{}  & 27.6 & 22.6  \\
\csref{} & \decs{}  & 21.1 & 13.2
\end{tabular}
\ifiwslt{\tabbleucaption}
%\ifinterspeech{\vspace{-2mm}}
\end{table}

\vspace{-2mm}
\subsection{Content Preservation}

To compare the difference in text simplification between machine translation and a human interpreter, we manually check the amount of information from the source text preserved in the translation. 
We employed two human annotators. They are both non-experts on the EP debates, non-native speakers of English, and native speakers of Czech. The first one, a professional translator, worked 5 hours and annotated 107 sentences. The second one, a computer linguist, contributed 20 sentences (1 hour). % and processed 20 sentences.
The annotators were provided with English revised transcripts of the whole document, and the translation candidates of automatic systems, interpreting and reference in Czech. They were all blinded and in random order. 
One random sentence from the source document was highlighted %and intended
for assessment.
The annotators were asked to express to what extent the information from the highlighted source sentence was preserved in the translation candidates, on a scale from 0 to 100. For comparability, they were asked to rate all the 6 candidates at once.
%, so we can compare the scores in \Cref{tab:annot}.

\Cref{tab:annot} indicates that \encs{} applied to the golden transcript preserves a similar amount of information as the manual translation. Involving any interpreter (\decs{} and \csint{}) leads to a considerable loss. ASR as the source for MT instead of gold transcripts significantly reduces translation quality, and loses further information (\enasr{}+\encs{} and \deasr{}+\decs{}).

The aggregated scores of the two annotators are consistent.
%The second annotator reported that he was able to guess whether the candidate translation was based on ASR, reference, or interpreting, with a high confidence. He noticed it from the translation quality and from sentence segmentation. 
%He also reports that in many cases, the amount
%with the least information preserved when applying ASR and MT to the output of the German interpreter (.2).
The second annotator reports that in many cases, the difference in non-ASR based translations were subtle and probably unimportant for the intended audience at the live event. For example, there was a substitution of ``president's office'' and ``the president'', as a subject in the sentence, and such cases were penalized slightly. In some cases, the translation of the highlighted sentence could not be found in the target, probably due to interpreter overload, and was largely penalized. It explains the low scores of the interpreting-based systems. Future evaluations could be provided by domain experts capable of considering the importance factor of particular facts. Also, the frequency of interpreting outages can be estimated by a targeted evaluation. 
%Although the system outputs were blinded, the sec

%We also notice the limits of our evaluation process. The source was
Our evaluation process has limitations, e.g. the source being
presented to the annotators only as English text, without audiovisual information. The gender of the speaker and addressed persons was thus often unclear, and its translation could not be evaluated. The interpreters use correct and consistent gender markers, while machine translation from English does not.

%We suggest that in the next work, the outages of interpreters should be evaluated. The content preservation should be also evaluated by experts who are aware of the audiovisual and background information, and can rate it
%The second limitation is that we neglect the real-time processing and latency. It is possible that 
%Or, an omission of one synonymic expression in conjunction, e.g.  
%if two similar entities or clauses were used in conjunction, only one was translated. 
%, were used in one phrase (e.g.  to each other, the second was omitted. Or, the translation used passive voice, which hides the 

%We note that our of evaluation neglected the effect of real-time processing. Manual interpreting has caused a considerable loss of information but this may have been necessary to ensure that the message remains comprehensible and easy to follow in the limited time.

% \XXX{M: Maybe we can show one cherry-picked sentence and compare how it is simplified by encs and decs}

\def\tabmipcaption{
        \caption{Manual assessment of information preserved.}
        \label{tab:annot}
        %\XXX{nechceme vykopnut stddev? nevyzera to presvedcivo}
        }
\begin{table}[t]
  \centering
  \ifinterspeech{\tabmipcaption{}\captiongap{}}
\begin{tabular}{l|@{~}r|@{~}r}
%\\
% cs-gold-interpreter	encs-mt-gold-short	decs-mt-gold-marian	cs-gold-reference	encs-mt-s2s-short	decs-mt-hybrid-marian
% 0.77±0.20	0.89±0.10	0.60±0.29	0.86±0.11	0.58±0.28	0.37±0.27

System                           & avg \p\ std  & avg \p\ std \\ \hline
\csref{}                         & 0.77 \p\ 0.32  & 0.86 \p\ 0.11 \\
%\textsc{EN}
EN src trans.+\encs{}     & 0.70 \p\ 0.33  & 0.89 \p\ 0.10 \\
\deint{} trans.+\decs{}     & 0.49 \p\ 0.37  & 0.60 \p\ 0.29 \\
\csint{}                         & 0.47 \p\ 0.39  & 0.77 \p\ 0.20 \\ 
\hline
\enasr{}+\encs{}                 & 0.38 \p\ 0.36  & 0.58 \p\ 0.28 \\
\deasr{}+\decs{}                 & 0.19 \p\ 0.29  & 0.37 \p\ 0.27 \\
\hline
Annotator                        & 107 sent., 5h & 20 sent., 1h \\
\end{tabular}
\ifiwslt{\tabmipcaption{}}
%\ifinterspeech{\vspace{-4mm}}
\end{table}

\section{Conclusion}

In this work, we release ESIC 1.0, a corpus with 10 hours of European Parliament speeches in English with transcripts, translations, and transcripts of simultaneous interpreting into Czech and German. We make it available for future work in speech translation and other areas:\\
\hbox to \hsize{\hss \url{http://hdl.handle.net/11234/1-3719} \hss}

We conclude that the automatic BLEU score is unable to distinguish whether the source-to-target or interpreter-to-target translation is better, due to the simplification feature of interpreting.
We compare direct and indirect speech translation by latency, and show that the indirect option could be comparable to relay interpreting. On the other hand, interpreter-based translation leads to shorter targets with significantly less complex vocabulary. A limited human assessment shows that more information is preserved in direct translation than in interpreting-based translations, and that far more content survives in translation from gold transcripts than from online ASR.% 
%
%We have performed experiments on this corpus in order to study the effect of human interpreters on the latency, quality and information content of the original speech and compared these to machine translation.
%The results showed that the human interpreters are currently faster than re-translating simultaneous machine translation. The 
%
%\section{Acknowledgements}
%
%\footnote{The research was partially supported by the grants 19-26934X  (NEUREM3)  of  the  Czech  Science Foundation, H2020-ICT-2018-2-825460 (ELITR) of the EU, 398120 of the Grant Agency of Charles University, and by SVV project number 260 575.}

\def\hidethis#1{}
\hidethis{\end{document}} % tohle zmate overleaf, že si nemyslí, že není ukončený begin-end document, takže nezačerveňuje všecko, a jinak to schválně nic nedělá

\bibliographystyle{IEEEtran}

\bibliography{references}

\end{document}